\pdfoutput=1
\documentclass[letterpaper, 10pt, conference]{ieeeconf}
\IEEEoverridecommandlockouts
\usepackage{cite}
\usepackage{amsmath,amssymb,amsfonts}
\usepackage{graphicx}
\usepackage{capt-of}
\usepackage{bm}
\usepackage{booktabs}
\usepackage{xcolor}
\usepackage{placeins}
\usepackage[colorlinks=true,allcolors=blue]{hyperref}

\graphicspath{{figures/}}

\newcommand{\modelname}{TrajectoryGPT}

\title{\LARGE \bf Whole-Body Humanoid Control via Stochastic Motion Tokenization}

\author{Michael Drolet$^{1}$, Oleg Arenz$^{1}$, and Jan Peters$^{1,2,3,4}$%
\thanks{$^{1}$Technical University of Darmstadt; $^{2}$German Research Center for AI (DFKI); $^{3}$hessian.AI; $^{4}$Robotics Institute Germany (RIG).
Corresponding author: Michael Drolet $\langle$michael.drolet@tu-darmstadt.de$\rangle$.}%
}

\begin{document}

\maketitle
\thispagestyle{empty}
\pagestyle{empty}

\begin{abstract}

While velocity-commanded RL policies are effective for controlling quadrupeds, they are a limited interface for humanoids that need to perform complex whole-body motions.
Furthermore, RL tracking policies for whole-body motions typically have limited information about the desired future motion, preventing them from performing controlled deviations to achieve long-term task success; such information is particularly relevant when a controller must prepare for transitions between behaviors.
We address these challenges using concise motion representations that can be generated by a high-level controller, given a behavior label or text command.
These representations can be fed directly into a tracking policy to provide a detailed motion reference.
Our method exhibits a high degree of flexibility in performing real-world robot motions such as dancing, fighting, or walking based on conditional information.
At matched capacity, our adopted stochastic motion tokenizer reconstructs motions from the Unitree G1 at a level that exceeds state-of-the-art approaches such as finite scalar quantization (FSQ) and the vector-quantized autoencoder (VQ-VAE).
The resulting discrete interface conditions whole-body tracking policies that execute generated dance, fight, and walk motions lasting $16.8$\,s each and a $35$\,s fight-to-dance sequence on a physical Unitree G1.
\end{abstract}

\section{Introduction}
Whole-body humanoid motion is high-dimensional and, at any instant, admits many plausible continuations: from mid-stride, a human may keep walking, turn, gesture, or transition into an entirely different behavior.
Classical humanoid control builds on dynamic models and optimization~\cite{raibert1986legged,kuindersma2016optimization,dicarlo2018dynamic} or motion primitives~\cite{schaal2006dmp}, while RL-based approaches~\cite{heess2017emergence,lee2020learning} can exhibit emergent behaviors but require reward specification, which is challenging to define explicitly for tasks such as human-like locomotion.
Controllers built around fixed gait generators or single-clip tracking fail to capture this multiplicity, ultimately reproducing only one trajectory or motion style well.
At the same time, methods based on maximum likelihood, such as behavioral cloning, require action labels via methods such as teleoperation or kinesthetic teaching, which is difficult to scale and infeasible for dynamic locomotion behaviors.
Furthermore, maximum likelihood is mode averaging, leading to muddied behaviors on unimodal policies.

A natural question is how to resolve the dilemma of choosing between a specialist policy that can perform one task very well and a generalist policy that can perform many tasks less effectively.
We address this problem by grouping motion into larger chunks and providing them as concise motion representations to a reinforcement learning policy, enabling it to focus on a particular motion that can be decided at a higher abstraction level.
This separation is inspired by motor learning theory~\cite{schmidt2014motorlearning}, which argues that, through practice, motor skills become grouped to abstract away nuances exhibited at each timestep.
This principle also appears in robot learning: Diffusion Policy and ACT both predict chunks of future actions rather than isolated commands~\cite{chi2023diffusion,zhao2023act}, leading to better performance than traditional imitation-learning methods~\cite{drolet2024}.
We therefore view it as a necessity to understand how to group motions in an unsupervised and efficient way, similar to the problem of tokenizing sentences in natural language processing.
Ideally, the abstracted representations should be i) compact for computational efficiency and generative utility, ii) seamlessly composable for long-horizon tasks, and iii) expressive enough to capture a wide range of sequences.
Our proposed framework is tailored to humanoid whole-body motion; however, the same key principles can be extended to other embodiments and problem settings.

\begin{figure}[t]
  \centering
  \includegraphics[width=\linewidth]{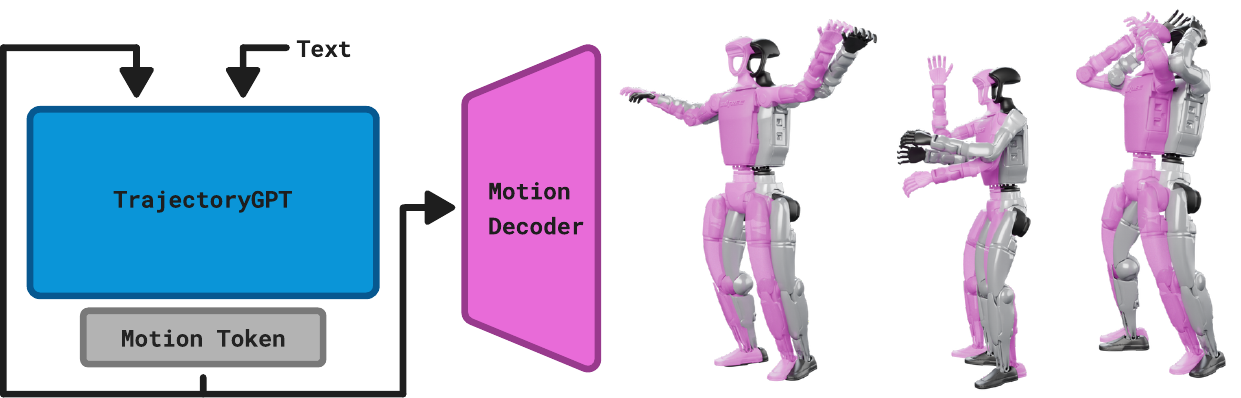}
  \caption{The Unitree G1 executing behaviors synthesized by our three-stage, next-token-prediction framework and tracked by a single policy.}
  \label{fig:teaser}
\end{figure}

To achieve these desirable characteristics, we use a motion tokenizer in the form of a discrete variational autoencoder (VAE).
Discrete VAEs offer high bit-efficiency compared to their continuous counterparts, and their categorical representations make them amenable to downstream generative tasks through the cross-entropy loss, which naturally leads to multimodal posterior sampling.
However, popular discrete tokenizers, such as FSQ and VQ-VAE, are either deterministic or limited to simplified distributions based on the mean-field assumption, which prevents them from correlating different tokens.
Furthermore, unlike their continuous counterparts, which aim at generative modeling, they are motivated from a compression perspective and are unable to control their entropy for optimizing the evidence lower bound (ELBO).
Instead, we adopt DAPS~\cite{daps} for the first stage of our pipeline, which is a stochastic discrete autoencoder whose encoder admits a categorical distribution that is factorized autoregressively and has controlled entropy.
When using DAPS to encode (or \emph{tokenize}) motions to discrete latents (or \emph{tokens}), we find that reconstruction is consistently better at matched bottleneck capacity.
Building upon this stage, we propose a second-stage model, \modelname{}, to generate a motion sequence autoregressively, and train it using the frozen VAE to encode prediction targets for autoregressive next-token prediction, as is common in related work~\cite{posegpt}.
\modelname{} enables us to schedule and sequence units of motion, all of which would be plausible under the distribution of trajectories in the training dataset.
Because \modelname{} is additionally conditioned on a context that describes the motion, it serves as an interface between the human and the robot.
This general interface can be extended beyond labels or text to include goal-conditioned information or vision input.
Finally, we realize these units of motion using a low-level tracking policy, as part of our third stage.
The tracking policy receives a buffer of latent codes---where tokens in the buffer can be set to special \emph{wildcard} tokens for transitions---to help inform the policy of the upcoming intended behavior.
In this regard, the policy can be seen as a decoder responsible for realizing the physical motion (under domain randomization for sim-to-real transfer).
In short, our contributions are:
\begin{itemize}
  \item A novel application of a stochastic motion tokenizer for the 29-DoF Unitree G1 with a matched comparison against existing state-of-the-art methods~\cite{fsq,vqvae}, across two motion datasets with multiple seeds.
  \item A context-conditioned \modelname{} that generates whole-body humanoid motion autoregressively, including from scratch, i.e., with no prior motion context.
  \item Deployment on a physical Unitree G1 (Fig.~\ref{fig:teaser}) of tracking policies trained entirely in simulation on a repertoire of behaviors sampled from our expressive model.
\end{itemize}

\section{Related Work}

\paragraph{Discrete tokenizers}
Discretizing data into tokens has been explored in both computer graphics and robotics.
In the case of computer graphics, discrete encodings are a popular way to distill a downstream model conditioned on rich context such as language for text-to-image generation~\cite{chang2022maskgit}.
For robotics, approaches such as BEAST~\cite{zhou2025beast} and FAST~\cite{pertsch2025fast}, based on splines and frequency-domain transformations, respectively, are popular ways to represent actions, or regression targets, for transformer-based behavioral cloning.
An alternative is to use learning-based approaches, where two of the most common methods for encoding discrete latents include the vector-quantized VAE (VQ-VAE)~\cite{vqvae} and finite scalar quantization (FSQ)~\cite{fsq}.
However, one drawback of vector-quantized methods is that decoding samples from a uniform latent prior (i.e., approximately sampling from $p(\bm x)$) often yields low-quality generative samples, as the decoder sees a limited combination of latent codes due to the compression objective~\cite{daps}.
Nevertheless, using discrete latents as tokens for motion synthesis (by reusing the VAE's decoder to convert tokens to kinematic sequences) is relatively nascent in robotics.
PoseGPT~\cite{posegpt} quantizes human poses with a VQ-VAE~\cite{vqvae} and uses the encoded codebook indices to represent the quantized motion tokens.
DAPS~\cite{daps} presents an alternative approach for training the encoder based on policy search; as a result, its latent space is a properly normalized, stochastic, categorical distribution, whose entropy and training stability can be carefully controlled.
Compared to vector-quantized and Gumbel-softmax-based approaches, DAPS has proven beneficial on benchmark-scale datasets such as ImageNet $256{\times}256$, in terms of both reconstruction loss and latent code utilization.
In this work, we adopt DAPS as a tokenizer for motion trajectories to take advantage of its benefits.

\paragraph{Physics-based motion imitation}
Motion imitation has been commonly achieved using reinforcement learning.
Adversarial methods such as AMP and ASE~\cite{amp,peng_ase_2022}, in which a discriminator-derived reward guides the policy toward expert-like behavior, have proven highly effective in simulation.
When framed as a standard RL problem with tracking rewards, DeepMimic~\cite{deepmimic} has gained much popularity.
These approaches address a key question for motion imitation: namely, given a desired kinematic sequence, which sequence of actions can generate the same behavior?
DeepMimic addresses this using an explicit reward signal, designed to penalize the policy for deviating from a reference motion (typically represented as joint angles and joint velocities~\cite{liao2025beyondmimic}) with a smooth exponential kernel.
In simulation, recent work has scaled motion imitation through simulated-avatar controllers~\cite{luo2023phc} and distilled universal latent motion representations~\cite{luo2024pulse}.
On real humanoid hardware, related advances include real-time whole-body teleoperation~\cite{he2024h2o,he2024omnih2o}, expressive body control with decoupled tracking~\cite{cheng2024exbody}, shadowing-based imitation~\cite{fu2024humanplus}, and large-scale motion tracking with SONIC~\cite{sonic}.
While these systems track observed or teleoperated motion (provided as FSQ encodings for SONIC), our method generates an indefinite sequence of motion from an autoregressive model.

Unsurprisingly, the utility of motion-tracking policies depends heavily on the reference motions.
A growing area of research investigates how to synthesize or retarget motions for trackers~\cite{omomo,omniretarget}.
An ongoing research problem, however, is improving the quality of motions when conditioned on task information such as goals or language.

\paragraph{Conditioned motion generation}
Non-physics-based, or ``kinematics only'', methods---largely popular in the computer graphics community---have become increasingly relevant for generating general-purpose motion for whole-body control.
Two dominant approaches, largely influenced by the adoption of multimodal generative models, are continuous diffusion and next-token prediction.
The motion diffusion model~\cite{tevet2023mdm} proposes a continuous diffusion-based approach for synthesizing motion from contextual information such as language.
CLoSD~\cite{tevet2025closd} closes the loop between such a text-to-motion diffusion planner and a physics-based tracking controller.
Autoregressive approaches include T2M-GPT~\cite{t2mgpt}, MotionGPT~\cite{motiongpt}, and the quadruped-focused T2QRM~\cite{t2qrm}.
TextOp~\cite{textop} recently demonstrated this generator--tracker separation for interactive, language-conditioned motion on humanoid hardware.
Recent large-scale efforts also learn generative models of motion, both kinematic~\cite{motionbricks} and physics-based~\cite{gpc}.

\section{Method}
\label{sec:method}

\begin{figure*}[!t]
  \centering
  \includegraphics[width=\textwidth]{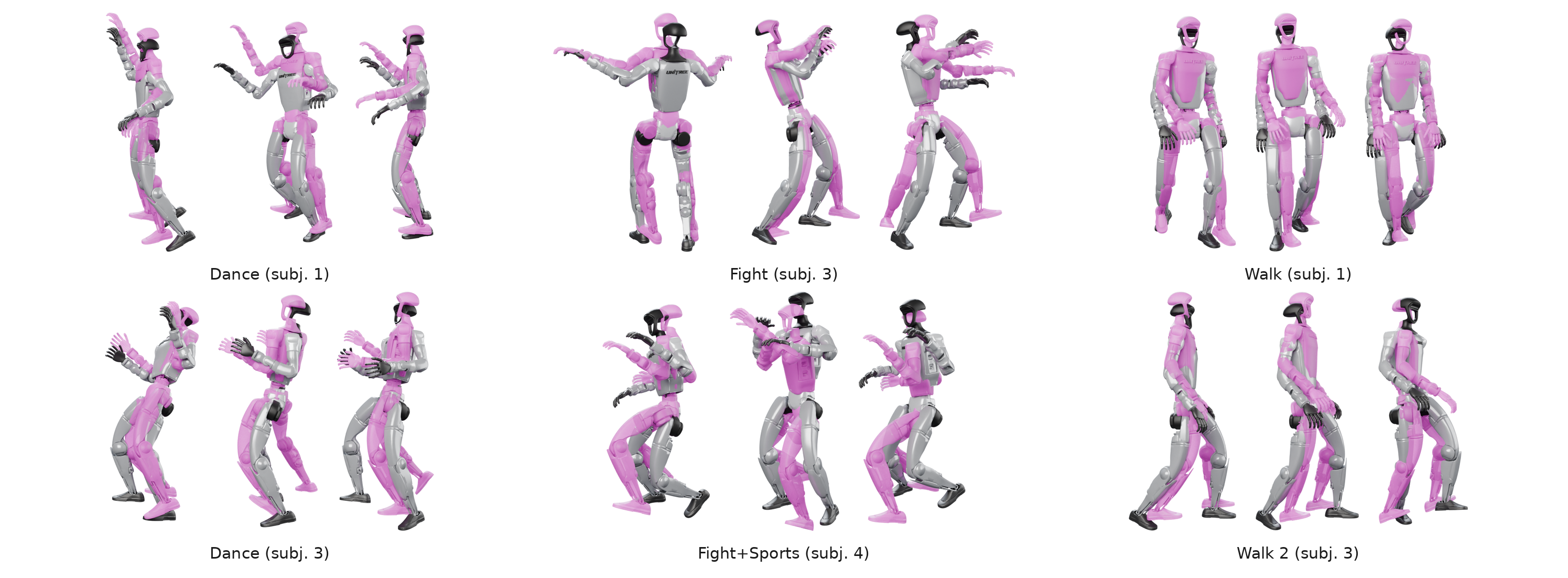}
  \caption{A repertoire of reference motions generated by the Stage~2 \modelname{} model (\emph{pink}) and tracked by the robot during Stage~3 training (\emph{silver robot}).
  The renderings are the true states of the robot and kinematics---a faithful depiction of the tracking behavior in the simulator.}
  \label{fig:gallery}
\end{figure*}
Our learning framework is divided into three stages: motion tokenization, autoregressive motion generation, and training of a low-level tracking policy.
We use LAFAN1~\cite{harvey2020robust} and the BONES-SEED dataset~\cite{bonesseed} to train the first stage, where the motions are already retargeted to the Unitree G1 29-DoF robot.
Motions are first resampled to $30$\,Hz and windowed into $24$-frame ($t{=}0.8$\,s) segments.
The robot state is the configuration $\bm q\in\mathbb{R}^{36}$ (root pose and 29 joint angles).

\subsection{Stage 1 --- Motion Tokenization (DAPS)}
The tokenizer encodes a derived $162$-dimensional feature vector $\bm x = [\bm q \,,\, \bm s]$ that augments $\bm q$ with 14 body-site positions and rot6d orientations $\bm s\in\mathbb{R}^{126}$ expressed relative to the root, computed from $\bm q$ by forward kinematics.
The Cartesian site terms allow the reconstruction objective to account for whole-body geometry rather than raw joint-angle error.
A recognition model $q_{\bm\theta}(\bm z\mid\bm x)$ encodes each window into $B{=}16$ latent tokens with vocabulary $V{=}1024$ ($160$ bits per window), and a generative model $p_{\bm\phi}(\bm x\mid\bm z)$ decodes them.
Both models are temporal transformers, and they jointly maximize the $\beta$-weighted ELBO
\begin{equation}
  \mathcal{L}^{\mathrm{ELBO}}_{\bm\theta,\bm\phi}=
  \mathbb{E}_{\bm x,\, \bm z\sim q_{\bm\theta}}
  \big[\log p_{\bm\phi}(\bm x \mid \bm z)\big]-\beta\,D_{\mathrm{KL}}\!\big(q_{\bm\theta}\,\|\,p\big)
  \label{eq:elbo}
\end{equation}
under a uniform prior.
The decoder first predicts the robot configuration $\bm q$ for all 24 frames.
This is then passed through differentiable forward kinematics to obtain the relative-site positions $\bm s$.
When evaluating the log-likelihood of $\bm x$, this design ensures that the joint and Cartesian components are consistent, while also constraining the output to be on the robot's kinematic manifold.

\subsubsection{Encoder update as KL-regularized policy search}
The encoder update is treated as a trust-region policy-search step based on REPS~\cite{peters2010reps,watson2023lbps}.
With per-sample return $R(\bm x,\bm z)=\log p_{\bm\phi}(\bm x\mid\bm z)$ and advantage $A = R - b(\bm x)$ against a Monte-Carlo baseline, each step solves
\begin{equation}
  \max_{q}\;\mathbb{E}_{\bm z\sim q}\!\left[A(\bm x,\bm z)\right]
  -\beta\,D_{\mathrm{KL}}\!\big(q\,\|\,p\big)
  \;\;\text{s.t.}\;\;
  D_{\mathrm{KL}}\!\big(q\,\|\,q_{\bm\theta}\big)\le\varepsilon .
  \label{eq:daps_ps}
\end{equation}
The Lagrangian of Eq.~\ref{eq:daps_ps} is maximized in closed form by
\begin{equation}
  q^{*}(\bm z\mid\bm x)\;\propto\;
  q_{\bm\theta}(\bm z\mid\bm x)^{\frac{\eta}{\eta+\beta}}
  \exp\!\Big(\tfrac{A(\bm x,\bm z)}{\eta+\beta}\Big),
  \label{eq:daps_qstar}
\end{equation}
where $\eta$ is the dual variable of the trust-region constraint.
Rather than solving the dual exactly, $\eta$ is adapted online so that the self-normalized importance weights $w \propto q^{*}/q_{\bm\theta}$ maintain a target effective sample size, which acts as a step-size controller in distribution space.
The new encoder is then the weighted maximum-likelihood projection of $q^{*}$ onto the model class, $\min_{\bm\theta} -\mathbb{E}[\,w\,\log q_{\bm\theta}(\bm z\mid\bm x)\,]$, such that no gradient flows through the autoregressive sampling and no straight-through estimator is needed.
The decoder is updated in parallel via maximum-likelihood estimation on decoded samples from the current parametric encoder.

\subsection{Stage 2 --- Autoregressive TrajectoryGPT}
With the tokenizer $q_{\bm \theta}$ frozen, an encoder--decoder transformer ($12$ layers, embed dim $384$) models the next token chunk given a context $\bm c$ containing the behavior label $\ell$ and the previous $N{=}8$ chunks of latent history.
Writing $q_{\bm\theta}=q_{\bm\theta}(\cdot\mid\bm x)$ and $q_{\bm\psi}=q_{\bm\psi}(\cdot\mid\bm c)$ for distributions over a complete token chunk, it minimizes the expected divergence $\mathbb{E}_{\bm x,\bm c}\big[D_{\mathrm{KL}}\!\big(q_{\bm\theta}\,\|\,q_{\bm\psi}\big)\big]$, which decomposes as
\begin{equation}
  \mathcal{J}(\bm\psi)
  =\mathbb{E}_{\bm x,\bm c}\big[\mathrm{CE}(q_{\bm\theta},q_{\bm\psi})-H(q_{\bm\theta})\big],
  \label{eq:ce}
\end{equation}
where $\mathrm{CE}(q_{\bm\theta},q_{\bm\psi})=-\mathbb{E}_{\bm z\sim q_{\bm\theta}}\log q_{\bm\psi}(\bm z\,|\,\bm c)$ and $q_{\bm\theta}$ is the frozen Stage-1 posterior at a chosen tokenization temperature, a hyperparameter in Fig.~\ref{fig:temperature}.
The context $\bm c = (\ell,\, \bm z_{\text{hist}})$ pairs the behavioral label $\ell$ (a one-hot class for LAFAN1 or a text embedding for BONES-SEED) with the $N{=}8$ historical latent chunks.
This loss maximizes the expected log-likelihood of \modelname{} under the encoder distribution $q_{\bm\theta}$ and, consistent with the VAE objective, includes the encoder entropy, which is constant in $\bm\psi$ and does not affect the gradient.
Since the cross-entropy term can be computed in closed form, the \modelname{} model fits our tokenizer's \emph{posterior}, not merely its mode.

We adopt a temperature hyperparameter $T$ on the logits to control the level of determinism when tokenizing; when small, the downstream models are encouraged to memorize the training data, and when large, the downstream models must generalize to more token sequences.
We find that low temperatures work well for capturing the motions in the original dataset.
During training, entries of the latent history $\bm z_{\text{hist}}$ are replaced with a \emph{start token} with small probability, so at inference the model can begin a rollout from the start token alone without requiring motion from the dataset.
Generation feeds each predicted chunk back as history, yielding open-ended label-conditioned rollouts.

\subsection{Stage 3 --- RL Tracking}
A single PPO~\cite{ppo} policy tracks the decoded references in the DeepMimic style: the reward compares the simulated state to the reference per feature group $d$ (joint positions and velocities; root-relative site positions, orientations, and velocities; and planar root velocity) as
\begin{equation}
  r_t=\sum_{d=1}^{D} w_d \exp\!\big(-\sigma_d\,(\tau^{d*}_t\ominus\tau^{d}_t)^2\big),
  \label{eq:reward}
\end{equation}
with no dedicated planar root-position term, so locomotion is propelled by physics rather than pinned to a global path.
The actor observes proprioception, a phase variable, and a rolling buffer of length $L{=}3$ latent chunks.
The first buffer slot encodes the current motion ($0.8$\,s) to mimic, and the next two encode the future motion to anticipate ($1.6$\,s).
This representation is bit-efficient: at $160$ bits per chunk, the buffer summarizes $2.4$\,s of intended motion in $480$ bits, rather than a short window of upcoming target frames in kinematic space.
Our hierarchy cleanly separates responsibilities: \modelname{} plans the motion given a context, and the tracking policy follows the motion plan.

Training uses $2048$ parallel GPU-simulated environments along with domain randomization for actuator gains, link masses, friction, and pushes.
Penalties on the joint accelerations and torques are also introduced to help reduce jitter on the policy and soften the robot's response for sim-to-real transfer.
All episodes start from a randomly chosen standing pose or from reference states.
When starting in a standing pose, the robot observes a wildcard latent chunk, indicating that it must learn to freely transition into a state that allows for good tracking performance on the reference motion.

\section{Experiments}
\label{sec:exp}
Our experiments assess each stage of the pipeline, culminating in deployment on the real robot.
We evaluate on the two datasets of Sec.~\ref{sec:method}: BONES-SEED and LAFAN1, across multiple seeds.
\mbox{Sec.~\ref{sec:exp:baseline}} compares tokenizers at matched bits, \mbox{Sec.~\ref{sec:exp:stage2}} evaluates conditional generation, and \mbox{Sec.~\ref{sec:exp:stage3}} explores how to provide latent codes to the low-level policy in simulation and on hardware.

\begin{figure*}[t]
  \centering
  \includegraphics[width=\linewidth]{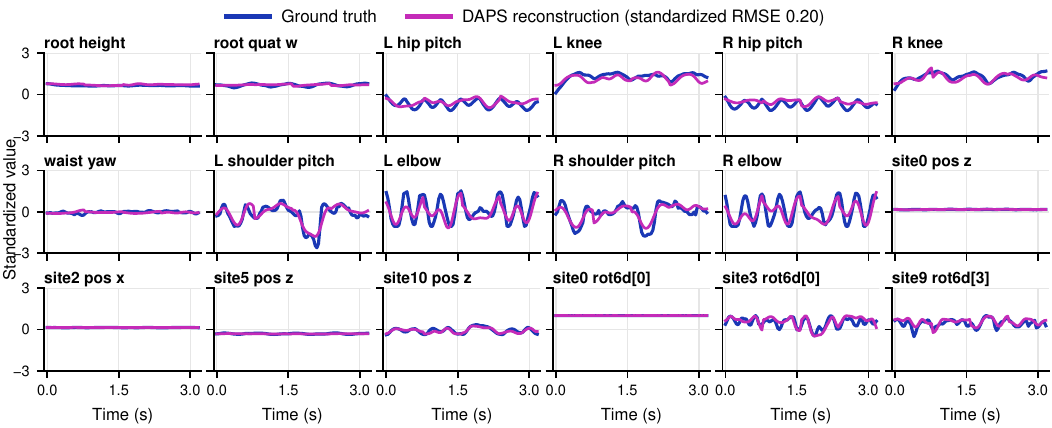}
  \caption{DAPS encode--decode on a G1 validation window: ground truth vs.\ reconstruction across representative channels of the standardized whole-body feature space (root state, joint angles, relative body-site positions and rotations) over $3.2$\,s of consecutive windows.
  The discrete code tracks the whole-body structure closely (overall standardized RMSE $\approx$0.20).}
  \label{fig:recon}
\end{figure*}

\subsection{Stage 1: Tokenizer Baseline at Matched Bits}
\label{sec:exp:baseline}
\begin{figure*}[t]
  \begin{minipage}[t]{0.48\textwidth}
    \centering
    \includegraphics[width=\linewidth]{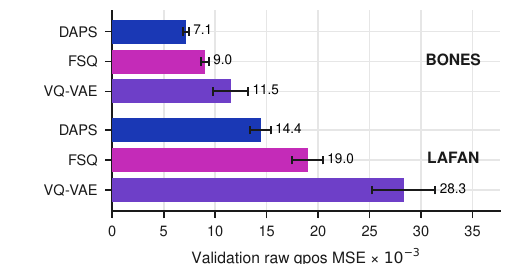}
    \captionof{figure}{Stage-1 validation joint position (qpos) MSE at a matched $160$-bit bottleneck, with mean$\pm$std over five seeds.}
    \label{fig:matchedbits}
  \end{minipage}\hfill
  \begin{minipage}[t]{0.48\textwidth}
    \centering
    \includegraphics[width=\linewidth]{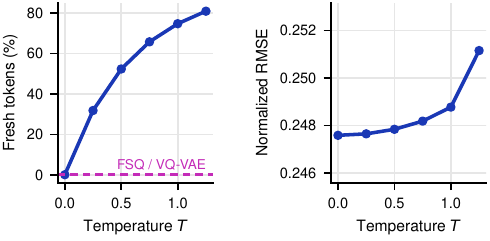}
    \captionof{figure}{The tokenize-temperature dial.
    \emph{Left:} fraction of tokens differing from the same segment's deterministic $T{=}0$ encoding.
    \emph{Right:} mean normalized reconstruction RMSE after decoding the sampled tokens.}
    \label{fig:temperature}
  \end{minipage}
\end{figure*}

All three tokenizers are trained at matched architectures and bottleneck: $16$ tokens, vocab size $2^{10}$, embed dim $128$, for $250$k steps across five seeds.
For FSQ, we use the recommended $[8,5,5,5]$ levels~\cite{fsq} for an approximately $2^{10}$ vocab size.
DAPS achieved the lowest validation joint position MSE on both BONES-SEED ($7.1 \pm 0.3$, versus $9.0 \pm 0.4$ for FSQ and $11.5 \pm 1.7$ for VQ-VAE) and LAFAN1 ($14.4 \pm 1.0$, versus $19.0 \pm 1.5$ and $28.3 \pm 3.0$, respectively), in units of $10^{-3}$.
DAPS also attained the highest validation conditional log-likelihood $\log p_{\bm\phi}(\bm x\mid\bm z)$ on BONES-SEED ($-910.4 \pm 1.3$, versus $-919.0 \pm 1.8$ and $-930.6 \pm 7.7$) and LAFAN1 ($-936.7 \pm 4.6$, versus $-949.3 \pm 4.0$ and $-994.7 \pm 16.7$), with all values reported as mean $\pm$ standard deviation over five seeds.
DAPS's advantage likewise held on the training splits, where it achieved the lowest raw-qpos MSE and highest conditional log-likelihood on both datasets.

Fig.~\ref{fig:temperature} demonstrates how DAPS exposes a controllable trade-off between token diversity and reconstruction fidelity.
Across all $5.58$ million segments in the BONES-SEED validation split, each segment was encoded at several temperatures and compared with its deterministic $T{=}0$ encoding.
The fraction of token positions that changed increased from $0\%$ at $T{=}0$ to $31.8\%$ at $T{=}0.25$, $52.4\%$ at $T{=}0.5$, $74.8\%$ at $T{=}1$, and $81.0\%$ at $T{=}1.25$.
Despite this substantial increase in latent diversity, normalized reconstruction RMSE increased only from $0.2476$ at $T{=}0$ to $0.2488$ at $T{=}1$ and $0.2512$ at $T{=}1.25$, corresponding to relative increases of approximately $0.5\%$ and $1.4\%$, respectively.
Thus, temperature provides a direct mechanism for supplying Stage~2 with diverse, valid tokenizations of the same motion while incurring only a modest reconstruction penalty.

\subsection{Stage 2: Conditional Generation}
\label{sec:exp:stage2}
Based on its superior reconstruction performance in Stage~1, we select DAPS as the tokenizer for the remainder of the pipeline.
We encode the training datasets with the frozen DAPS encoder and train \modelname{} to predict the next DAPS token chunk from the behavior condition and latent history.
At inference, generation begins from either an encoded motion prefix or the learned start token and proceeds one chunk at a time.
Each predicted chunk is decoded into $0.8$\,s of G1 kinematics and returned to the latent context, producing open-ended, context-conditioned references for the Stage~3 tracking policy.
While cross-entropy serves as the Stage~2 training objective, a lower loss indicates increasingly accurate token prediction within the context of known continuations, and a value approaching zero can be consistent with memorization.
We therefore use Stage~2 to produce long reference motions for Stage~3 and do not treat cross-entropy alone as a diagnostic for motion quality.
Human-preference metrics could assess these generated motions and could also replace supervised learning with RL-based optimization in Stage~2.

\subsection{Stage 3: Tracking and Transfer}
\label{sec:exp:stage3}
We train three latent-conditioned PPO policies with the reward and observations of Sec.~\ref{sec:method} to test the same learned interface with different motion-bank compositions.
One policy tracks $25$ diverse generated motions, another tracks three generated base motions and all nine transition pairings (including same-style transitions), and a third tracks a smaller bank of five motions.
Stage~3 uses a subset of the motions produced by the LAFAN1-trained \modelname{} model; Fig.~\ref{fig:gallery} shows six examples from the $25$-motion policy.

\begin{figure*}[t]
  \centering
  \includegraphics[width=\textwidth]{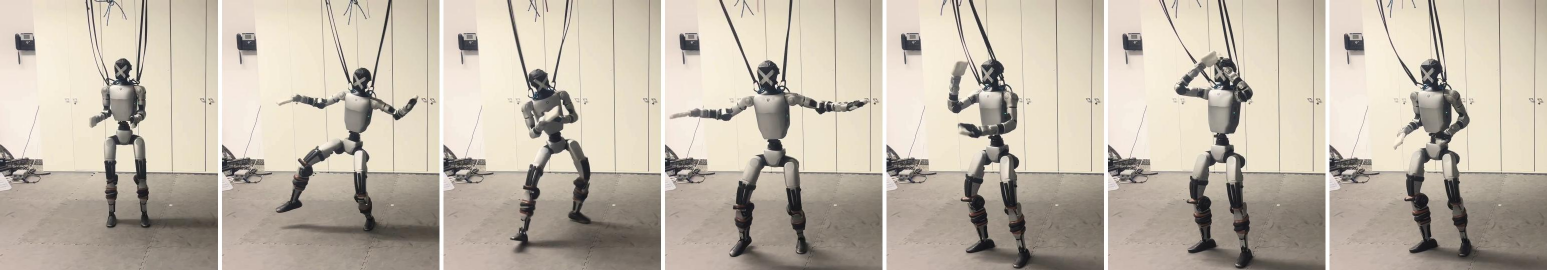}
  \caption{Real-robot behavior switch on the Unitree G1: a single policy transitions from a fight motion into a dance through a wildcard bridge, during which no reference is prescribed, completing the full $35$\,s sequence upright.}
  \label{fig:realrobot}
\end{figure*}

Each stitched sequence concatenates two independently generated motions: $21$ chunks ($16.8$\,s) from the first behavior, two wildcard chunks ($1.6$\,s) with no prescribed kinematic reference, and $21$ chunks from the second behavior, for a total duration of $35.2$\,s.
On the G1, the five-motion policy completed generated dance, fight, and walk motions, each lasting $16.8$\,s, at $50$\,Hz while remaining upright and advancing through the complete latent sequence (Fig.~\ref{fig:hardwaretracking}).
The stitching policy completed the full fight-to-dance sequence, including the wildcard interval (Fig.~\ref{fig:realrobot}).

\subsection{Implementation}
All stages run in JAX with XLA acceleration.
Each stage trains on a single GPU having $24$--$80$\,GB of VRAM; Stage~3 simulates $2048$ MuJoCo environments on one GPU.
We generate the reported Stage~3 reference banks offline, so \modelname{} latency is excluded from the tracking measurements.
Deployment consumes the resulting precomputed latent trajectories.
The stages are trained sequentially, and each is frozen before the next begins.

\subsection{Hyperparameters}
Table~\ref{tab:hparams} presents the pipeline hyperparameters for both datasets.
The DAPS ESS target transfers without modification between BONES-SEED and LAFAN1; $\beta$ is annealed from $0.3$ to $10^{-2}$ and $3{\times}10^{-3}$, respectively.
The desired KL ($0.01$--$0.02$) is set per Stage-3 policy.
For hardware transfer, we tune the action-rate penalty to reduce jitter without excessively damping motions the policy has to track.

\begin{figure}[t]
  \centering
  \includegraphics[width=\columnwidth]{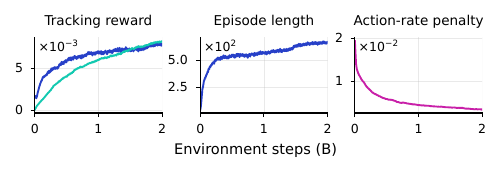}
  \caption{Stage-3 PPO training of the deployed stitching policy (Fig.~\ref{fig:realrobot}).
  \emph{Left}: Tracking reward for joint position (\emph{blue}) and joint velocity (\emph{teal}).
  \emph{Middle}: Average episode length of all environments throughout training.
  \emph{Right}: Action-rate penalty, as part of the total environment reward.
  Training statistics are shown over $2$B environment steps.}
  \label{fig:ppocurves}
\end{figure}

\begin{table}[t]
  \vspace{3mm}
  \centering
  \caption{Pipeline hyperparameters.}
  \label{tab:hparams}
  \begin{tabular}{@{}lll@{}}
    \toprule
    Stage & Parameter & Value \\
    \midrule
    Data    & rate / window            & $30$\,Hz / $24$ frames \\
    Data    & feature dim              & $162$ ($36$ qpos $+126$ rel.\ sites) \\
    Stage 1 & tokens $B$ / vocab $V$   & $16$ / $1024$ ($160$ bits) \\
    Stage 1 & embed / batch / steps    & $128$ / $256$ / $250$k \\
    Stage 1 & $\beta$ sched.\ / ESS / MC & $0.3\!\to\!10^{-2}$--$3{\times}10^{-3}$ / $0.5$ / $16$ \\
    Stage 2 & layers / embed / history & $12$ / $384$ / $N{=}8$ chunks \\
    Stage 3 & PPO envs / rollout / $\gamma$ & $2048$ / $50$ / $0.99$ \\
    Stage 3 & desired KL / GAE $\lambda$     & $0.01$--$0.02$ / $0.95$ \\
    Stage 3 & $w_d$ qpos / qvel             & $2$ / $1$ \\
    Stage 3 & $w_d$ sites / rquat / rvel    & $4.5$ / $2$ / $0.5$ \\
    Stage 3 & $w_d$ planar root vel         & $0.2$ \\
    Stage 3 & action-rate / latent buffer    & $-0.4$ / $3$ chunks \\
    Stage 3 & total steps                    & $2$B \\
    \bottomrule
  \end{tabular}
\end{table}

\newpage
\section{Discussion}
Using the same discrete representation for motion generation and tracking provides a direct interface between \modelname{} and the low-level policy.
\modelname{} specifies the intended motion at the token level, while the tracking policy realizes the decoded motion under the robot's dynamics.
This separation lets us assess errors introduced by reconstruction and physical execution, while errors in autoregression are more difficult to quantify due to the absence of a ``ground truth''.
DAPS can assign different token sequences to the same motion, and Fig.~\ref{fig:temperature} shows that many retain similar reconstruction accuracy.
By changing $T$, \modelname{} can be trained on the most likely encoding of each window or on sampled alternatives, without changing either the tokenizer or the Stage~2 architecture.

Beyond compression, the latent interface determines how much future intent the controller can observe.
Each $0.8$\,s chunk carries $160$ bits, or $200$ bits/s, and the three-chunk buffer exposes $2.4$\,s of motion while keeping raw reference poses out of the actor's observation space.
The present \modelname{} is not conditioned on the low-level robot state, so this look-ahead provides a \emph{preplanned} rather than a \emph{re-planned} motion.
To close the loop between the two levels~\cite{tevet2025closd}, \modelname{} could instead condition online on the robot's own motion: re-encoding the measured history with the frozen DAPS encoder and using these tokens in place of the previously generated chunks requires no change to the architecture, since both share the same token space.
Because \modelname{} is trained on dataset histories, however, the re-encoded robot motion could be out of distribution.
Relatedly, three variables remain interesting to explore: the chunk duration ($t{=}0.8$\,s) in Stage~1, the length of the latent history ($N{=}8$) in Stage~2, and the length of the latent buffer ($L{=}3$) in Stage~3.
Because chunk duration induces different reconstruction errors for a fixed bottleneck, it directly affects the reconstruction performance.
Chunk duration may also impact the smoothness of the generated motion for Stage~2, since longer contexts can shrink the set of plausible future continuations.
The same applies to $N$, where larger values can also lead to overfitting.

Wildcard chunks remove the reference-tracking objective for the transition interval.
The regularization penalties stay active to encourage the robot to remain stable.
This gives the state-conditioned policy freedom to connect the surrounding motions while the future latent chunks remain visible.
The fight-to-dance trial shows that such a transition can be executed on hardware.
Supplying a prior motion for the transition interval could further shape this behavior.

\begin{figure*}[t]
  \centering
  \includegraphics[width=0.98\textwidth]{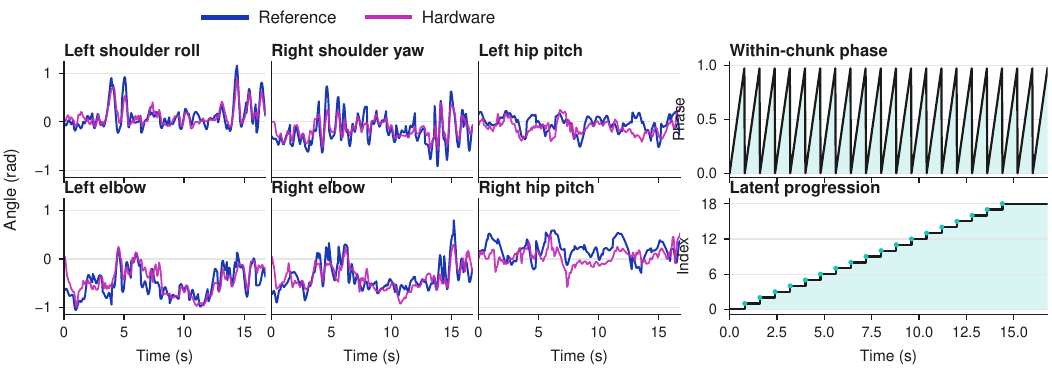}
  \caption{Real-robot execution of a generated dance on the Unitree G1.
  Six joint angles from the measured hardware state (\emph{magenta}) are compared with the source kinematics (\emph{blue}), linearly resampled from $30$ to $50$\,Hz.
  The rightmost column shows within-chunk phase and latent-chunk progression.}
  \label{fig:hardwaretracking}
\end{figure*}

A natural next step is a scaling study that varies the repertoire size and transition difficulty and reports per-motion tracking performance.
An additional Stage~3 test could hold out latent buffer sequences during policy training and introduce them only at evaluation.
This would distinguish the reuse of motion skills from the memorization of complete latent schedules seen in training.

Fig.~\ref{fig:hardwaretracking} compares the measured robot state with the source kinematics.
The desired joint configuration is not provided to the deployed policy, which receives proprioception, phase, and the rolling latent buffer.
Error residuals include both policy-tracking and sim-to-real error.
Because the reference banks are generated offline and held fixed, these residuals arise downstream of \modelname{}.
Neither \modelname{} nor the DAPS decoder runs in the $50$\,Hz deployment loop.
Because the actor consumes tokens directly, high-level updates are needed only when a new $0.8$\,s chunk is required.
With two future chunks already buffered, an online \modelname{} could run asynchronously and refresh the plan before the $1.6$\,s look-ahead is exhausted.
Online generation is primarily outside the scope of our algorithmic contribution and was not needed to evaluate the method, but could be added if required.

Each $24$-frame source chunk is played back over its original $0.8$\,s duration at $50$\,Hz.
The phase observation is local to each chunk and is advanced at a fixed rate.
Fig.~\ref{fig:hardwaretracking} (right) shows this phase resetting at every chunk boundary while the latent index advances through the sequence.
Adapting the phase to contact or tracking progress could relax the imposed timing.
Furthermore, because planar root position is not explicitly tracked, the tokens specify motion style without specifying a global path.
Adding a goal or path-conditioned context would allow this work to be extended to areas such as navigation, where the motion must also follow a commanded route.

\newpage
\section{Conclusion}
We presented a three-stage pipeline in which a discrete autoencoder provides the shared interface between motion generation and whole-body control.
At a matched $160$-bit bottleneck, DAPS reconstructs G1 motion more accurately than FSQ and VQ-VAE while retaining temperature-controlled alternative encodings.
\modelname{} learns sequences in this compact space, and a latent-conditioned policy tracks their decoded motions without receiving raw reference poses.
This division allows the high-level model to organize behavior at the token level while the low-level policy remains reactive to environment dynamics.

Our results suggest that the recipe behind GPT-style language models can carry over to humanoid control: a stochastic tokenizer defines a motion vocabulary, a transformer composes behavior one token at a time, and a single tracking policy reads the token stream on the robot.
On a physical Unitree G1, the policy executed individual generated motions and a full fight-to-dance sequence.
The measured-state comparison further shows that the deployed policy follows the intended source motion without the motion being exposed explicitly to the actor.
Together, these results suggest how discrete motion tokens can serve as both a compact generative vocabulary and a deployable interface for humanoid control.

\vfill

\section*{Acknowledgements}
This research has been partially supported by the German Federal Ministry of Research, Technology and Space (BMFTR) within the project RobInTime (Grant: 01IS25002B).
This project has received funding from the European Union’s Horizon Europe research and innovation program (Grant agreement ID: 101298421).
This project has been supported by a hardware donation by NVIDIA through the Academic Grant Program.
We gratefully acknowledge support from the hessian.AI Service Center (funded by the Federal Ministry of Education and Research, BMBF) and the hessian.AI Innovation Lab (funded by the Hessian Ministry for Digital Strategy and Innovation).
The authors used Claude (Anthropic) and ChatGPT (OpenAI) to assist with language editing and code revision.
All such content was reviewed and verified by the authors, who take full responsibility for the paper.

\bibliographystyle{IEEEtran}
\bibliography{refs}

\end{document}